\documentclass[review]{elsarticle}
\graphicspath{ {./figures/} }
\usepackage{hyperref}
\usepackage{float}
\usepackage{verbatim} %comments
\usepackage{apalike}
\restylefloat{figure}
\restylefloat{table}
\usepackage{amsfonts}
\usepackage{amsmath}
\usepackage{lipsum}
\usepackage{mwe}
\usepackage{atbegshi,multido,graphicx}

\makeatletter
\newcommand{\removepage}[1]{%
  \def\@drop@this@page{\AtBeginShipoutDiscard}%
  \multido{\iA=1+1}{#1}{%
    \expandafter\gdef\expandafter\@drop@this@page\expandafter%
      {\expandafter\AtBeginShipoutNext\expandafter{\@drop@this@page}}%
    }%
  \@drop@this@page%
}

\journal{Expert Systems with Applications}

\biboptions{authoryear}

\begin{document}
\removepage{1}%
\begin{frontmatter}

%\begin{titlepage}
\begin{center}
\vspace*{1cm}

\textbf{ \large Feature Importance in Gradient Boosting Trees with Cross-Validation Feature Selection}

\vspace{1.5cm}

% Author names and affiliations
Afek Ilay Adler (afekadler@mail.tau.ac.il) and Amichai Painsky (amichaip@tauex.tau.ac.il)\\

\hspace{10pt}

\begin{flushleft}
\small  
$^a$ The Industrial Engineering Department, Tel Aviv University, Ramat Aviv, Israel

\begin{comment}
Clearly indicate who will handle correspondence at all stages of refereeing and publication, also post-publication. Ensure that phone numbers (with country and area code) are provided in addition to the e-mail address and the complete postal address. Contact details must be kept up to date by the corresponding author.
\end{comment}

\vspace{1cm}
\textbf{Corresponding Author:} \\
Last Author \\
Chaim Levanon St 30, Tel Aviv, Israel
\\
Tel: (972) 50-868-1368 \\
Email: amichaip@tauex.tau.ac.il

\end{flushleft}        
\end{center}
%\end{titlepage}

\title{Feature Importance in Gradient Boosting Trees with Cross-Validation Feature Selection}

\author{Afek Ilay Adler}
\ead{afekadler@mail.tau.ac.il}

\author{Amichai Painsky}
\ead{amichaip@tauex.tau.ac.il}

\address{The Industrial Engineering Department, Tel Aviv University, Israel}

\begin{abstract}
Gradient Boosting Machines (GBM) are among the go-to algorithms on tabular data, which produce state of the art results in many prediction tasks. Despite its popularity, the GBM framework suffers from a fundamental flaw in its base learners. Specifically, most implementations utilize decision trees that are typically biased towards categorical variables with large cardinalities. The effect of this bias was extensively studied over the years, mostly in terms of predictive performance. In this work, we extend the scope and study the effect of biased base learners on GBM feature importance (FI) measures. 
We show that although these implementation demonstrate highly competitive predictive performance, they still, surprisingly, suffer from bias in FI. By utilizing cross-validated (CV) unbiased base learners, we fix this flaw at a relatively low computational cost. We demonstrate the suggested framework in a variety of synthetic and real-world setups, showing a significant improvement in all GBM FI measures while maintaining relatively the same level of prediction accuracy.
\end{abstract}


\end{frontmatter}

\section{Introduction}
In recent years, machine learning (ML) has gained much popularity and became an integral part of our daily life. Current state-of-the-art algorithms are complex models which are difficult to interpret and rely on thousands, and even billions of parameters. Gaining insight into how these complex algorithms work is a key step towards better understanding our environment. Further, it is a crucial step for enabling ML algorithms to support and assist human decision making in complex fields (for example, as in medicine \citep{lundberg2018explainable}). FI is one of the basic tools for this purpose. The general FI framework scores input variables by their contribution to the predictive model, allowing us to gain insight into which features are important for this task.
GBM \citep{friedman2001greedy} is an ensemble of base (weak) learners. These learners are typically standard implementations of tree-based models such are CART \citep{breiman1984classification}, C4.5 \citep{quinlan1986induction}, and others. GBM's are among the current state-of-the-art ML techniques on tabular data in a variety of tasks such as click prediction \citep{richardson2007predicting},  ranking \citep{burges2010ranknet} and others. Besides its accuracy, the GBM framework holds many virtues which makes it a favorable choice for many learning tasks. It is efficient, can handle categorical variables and deals with any kind of differential loss function. Further, it supports missing values and invariant to feature scaling. However, it is well known that GBM base learners are biased towards categorical variables with large cardinalities \citep{breiman1984classification}.   As a consequence, ensemble methods that rely on such biased learners suffer from the same flaw \citep{strobl2007bias}. %Common unbiased partitioning methods depend on a-priori modeling assumptions \citep{loh2002regression}, or computationally expensive \citep{hothorn2006unbiased}.

In this work we study the effect of base learners' bias on FI measures, in standard GBM implementations. We show that in the presence of categorical features with large alphabets, most FI measures are typically biased. This results in miss-interpreted models. To overcome this caveat, we utilize a cross-validated boosting (CVB) framework. CVB addresses the tree's bias through a simple modification: during the training stage of the tree,  a variable is selected for splitting based on its cross-validated performance, rather than on its training sample performance (see \cite{painsky2016cross}, for example). 
This way, we conduct a ''fair” comparison between features, utilizing only high cardinality categorical features that show to be informative, in terms of their estimated generalization abilities. 

 The rest of this manuscript is organized as follows, in Section \ref{Background} we review previous related work. In Section \ref{Formulation Of CVB} we formalize and define cross-validated boosting (CVB). Section \ref{Methods} outlines the experiment design in which we compare CVB to other GBM frameworks. In Section \ref{simulation_studies} we compare CVB to common open-source GBM implementations and show that CVB is the only algorithm that does not suffer from biased FI in high cardinality features through simulation studies. Finally, Section \ref{Real Data Case Studies} contains two comprehensive real data case studies, showing that CVB FI is more reliable than alternative schemes. We conclude the manuscript in Section \ref{discussion}. 
 
\section{Background}
\label{Background}
\subsection{Decision Trees}
Decision trees are a long standing staple of predictive modeling. Popular tree building algorithms can handle both numerical and categorical variables, and build models for regression, binary and multi-class classification. Decision trees are built recursively to minimize a given loss function. At each step, the split which optimizes a given criterion is selected from all possible splits over on all variables.
We hereby present the CART approach, which is one of the most widely used tree implementation in current GBM implementations.

Let $\{x_i, y_i\}_{i=1}^{N}$ be a training set with $N$ samples , where $x_i \in \mathbb{R}^p, y_i \in \mathbb{R}$.
% Assume, without loss of generality, that the variables $x_1, ..., x_q$  are categorical with
% number of categories $K_1, ..., K_q$ respectively, and the features $x_{q+1}, ..., x_{p},$
% are numerical or ordinal.
The standard CART approach is a binary tree which utilizes either squared error loss impurity criterion (regression) or the Gini index (classification) to evaluate a possible split.  At each internal node of a tree, a split $s$ is a partition of $n$ observations to two disjoint subsets $R(s)$ and $L(s)$. 
In regression tress, the squared error impurity criterion is defined as
\begin{equation} \label{equation1}
\mathcal{I}(s)  = \Sigma_{i \in L(s)} (y_i - \bar{y}_L)^2 +\Sigma_{i \in R(s)} (y_i - \bar{y}_R)^2,
\end{equation}
where $\bar{y}_L, \bar{y}_R$ are the means of the response $y$ over the sets $L(s), R(s)$ respectively. The split gain (SG) is defined as the amount of decrease in impurity of the split:
\begin{equation} \label{equation2}
SG(s)  =  \Sigma_{i=1}^n (y_i - \bar{y})^2 - \mathcal{I}(s).
\end{equation}

At each node, CART examines all possible splits for each feature and selects the split with the minimum impurity (maximum gain). For each variable $j$, we denote the set of possible splits by $S_j$ and their cardinality by $|S_j|$. 
Let  $K_j$ be the number of categories in a categorical feature $j$.
For each numerical/ordinal variable, CART considers only splits along the sorted variable. Therefore, there are $|S_j| \leq n-1$ possible splits between its unique sorted values. For a categorical variable, there are $|S_j| = 2^{K_j -1}$ possible binary splits, but it is sufficient to sort the categories by their mean response value and search for splits along this axis \citep{breiman1984classification}, reducing the complexity to only $K_j-1$ splits.  

Without any regularization mechanism, terminal nodes are split and deep, complex trees are formed. Thus, stopping criteria are typically applied to prevent over-fitting and reduce tree complexity. Several common stopping criteria are the maximum tree depth, minimum samples to split, minimum samples leaf and minimum impurity decrease. It is also possible to grow a tree to its maximum depth and then to prune it, such as in minimal cost-complexity pruning \citep{breiman1984classification}. 

An additional commonly used decision tree implementation is C4.5 \citep{quinlan1986induction}.
C4.5 considers K-way splits for categorical features (as opposed to binary splits in CART). C4.5 trees may be problematic for large cardinality categorical features as they result in splits with very few samples, and henceforth induce variance.
Further, C4.5 trees are not used in recent GBM implementations. 
A comprehensive discussion on different splitting frameworks is provided in \citep{loh2014fifty}.

\subsection{Gradient Boosting Trees}
GBM is an ensemble model of $M$ regression trees, trained sequentially one after the other, in order to predict the previous trees' errors. 
GBM may be viewed as gradient descent in a functional space, where in each iteration of GBM, a tree is grown to estimate the gradient of the objective function.
Let $F_m : \mathbb{R}^p \rightarrow \mathbb{R}$ be a predictive model at stage $m$, where $m = 1,....,M$. Let $\mathcal{L}(y_i,F_m(x_i))$ be a differentiable loss function. For example, $\mathcal{L}_{SE}(y_i,F_m(x_i))=(y_i-F_m(x_i))^2$ is the squared error loss in the standard regression case.

GBM initializes with a constant model by estimating a parameter that minimizes the overall loss (for the squared error $F_0(x_i) = \bar{y}$). Then, at every iteration $m$, a regression decision tree $h_m$ is trained to minimize the empirical risk
\begin{equation*} 
h_m = \operatorname*{argmin}_h \Sigma_{i=1}^N  \mathcal{L}(y_i,F_{m-1}(x_i) + h(x_i)).
\end{equation*}
$h_m$ is obtained by fitting a regression tree to the gradients of each sample with respect to the current estimator at stage $m$. 
The optimal step size is calculated per each leaf by a line search. $h_m$ can be seen as the best greedy step in order to minimize the empirical risk. In order to reduce over-fitting, a learning-rate $\alpha$ is used and the model is updated, $F_m =  F_{m-1} + \alpha h_m$. This process is repeated until $M$ trees are grown. 
Later, inspired by Breiman's bootstrap-aggregating (bagging) \citep{breiman1996bagging}, Friedman showed that a stochastic modification to GBM  can substantially improve the prediction error of GBM \citep{friedman2002stochastic}. In its stochastic form, GBM fits each tree on a sub-sample of the training-set, sampled at random without replacement. 

\subsection{GBM Implementations} \label{gbm_implementations}
There are many implementations of GBM in the literature. Many of these implementations follow Friedman's GBM,  which is based on CART trees.  
In the age of big data, modern ML tasks require very large datasets. Therefore, the need for more scalable, distributed, GBM solutions arouse (such as  \cite{tyree2011parallel}).
This line of work mostly focuses on efficient implementation of GBM in distributed environments such as MapReduce \citep{condie2010mapreduce}. Here, the most notable implementation is XGBoost \citep{chen2016xgboost}. XGBoost has two main advantages. First, it unifies previous ideas of distributed computing with novel optimizations such as out-of-core-computation and sparsity-awareness. Thus, producing much lower run-time both on a single machine and in a distributed environment. Second, XGBoost introduces a new regularized learning objective which penalizes the complexity of the model.
The result is a highly scaled end-to-end open-source implementation that produces state-of-the-art results, winning multiple data science open competitions. 

Unfortunately, the main drawback of the XGBoost framework is the lack of categorical variables support.
Specifically, in order to utilize a categorical feature one needs to transform it to numeric variables such as in one-hot-encoding or to use \textit{term statistics}. One-hot-encoding is a valid method to transform categorical features to numeric ones, but in cases of high cardinality variables, it substantially increases the number of features and henceforth the execution time. \textit{Term statistics} is a method to transform a categorical feature to numeric by using the target value. Here, each category is replaced with its respective mean response value or a smoothed version of it \citep{prokhorenkova2018catboost}. The main problem with this approach is that it implies an order among the categories, and limits the expressive
power of CART in categorical splits.

Following XGBoost, Microsoft introduced LightGBM (LGBM) \citep{ke2017lightgbm}. LGBM is a faster implementation of GBM which supports categorical features. LGBM uses sampling techniques to exclude data instances with small gradients and \textit{exclusive feature bundling} to reduce the number of features tested to split at each node while achieving almost the same accuracy. More importantly, LGBM supports categorical features out-of-the-box and follows the original CART approach. Lately, CatBoost pushed the envelope of treating categorical features in GBM even further. CatBoost
addresses a \textit{target leakage} (see \cite{prokhorenkova2018catboost}) to significantly improve the prediction error.  
For categorical variables, CatBoost uses ordered
\textit{term statistics} and it is henceforth more memory efficient than LGBM.
Further, CatBoost utilizes oblivious trees as base learners. Oblivious decision trees are decision trees for which all nodes at the same level use the same feature to split. These trees are balanced, less prone to over-fitting, and allow speeding up execution at testing time. Finally, CatBoost uses combinations of categorical features as additional categorical features to capture high-order dependencies between categorical features. 

\subsection{FI Methods}
Understanding the decision-making mechanism of a ML algorithm is an important task in many applications.
First, in several setups, attaining the best prediction is not our sole intent; we would like to know more about the problem and the data. Second, model interpretability is a key step for achieving and quantifying other desirable properties in artificial intelligence (AI) such as fairness, robustness, and causality \citep{doshi2017towards}. Finally, as AI enters new domains as health and transportation (autonomous vehicles), interpretation is essential to trust those mechanisms and adopt them as an integral part of our lives. 

While some ML algorithms are intrinsically interpretive (linear regression), they tend to be over-simplistic and generally suffer from a high bias. As modern ML evolves and computing power increases, complex algorithms are developed to attain superior prediction capabilities but in turn provide little interpretation. For this reason, many methods are developed to make these non-interpretable ML methods more understandable. These methods typically provide summary statistics for each variable, which can also be visualised for better understanding. 

Global FI measures attempt to summarise to which extent each feature is important for the prediction task. Given an algorithm, train and/or test data, a global FI method outputs a single number per each feature that reflects its importance.
A local FI measure outputs a FI vector per each observation. Thus, it provides a more detailed, personalized FI. Local explanations are crucial in fields such as personalized medicine. For instance, local explanations can assist anesthesiologists to avoid hypoxemia during surgery, by providing real-time interpretable risks and contributing factors \citep{lundberg2018explainable}.
Interaction FI are methods that measure the strength of the interaction between two features.
A FI method is said to be model agnostic if it can be applied in the same manner for any ML algorithm. On the other hand, a FI measure is said to be model specific if it only applies for a specific algorithm \citep{molnar2020interpretable}. 

While FI aims to understand which features are important given a trained model, feature selection methods usually train a model several times in order to reduce dimensionality for speed considerations \citep{pan2009feature} or to avoid over-fitting in cases such as ''large-p small-n" tasks. While FI measures can be used as a means for variable selection by taking the most influential features, variable selection methods can not always be used as FI measures. For example, methods that rely on greedy backward elimination of features (such as  \citep{kursa2010feature,gregorutti2017correlation}), still need to use a FI measure to score each variable in the final selected feature subset.

\subsubsection{Global FI Methods}
As mentioned above, global FI measures can be agnostic or model-specific. A global model-specific FI for GBM can be derived
by evaluating its individual trees FI and averaging them across all trees in the ensemble \citep{friedman2001greedy}. 
Given a tree, there are many methods to interpret how important is a feature by inspecting the non-terminal nodes and their corresponding features which were used for splitting. The most prominent measures are:   
\begin{itemize}
  \item Split count - the number of times a feature is selected for a split in a tree.
  \item Cover - the number of observations in the training set that a node had split.
  Specifically, for every feature $j$, Cover is the total number of observations in splits that involve this feature.
  \item Gain or the decrease in impurity \citep{breiman1984classification} - for every feature $j$, Gain is the sum of SG (\ref{equation2}) that utilize this feature. 
\end{itemize}

These measures are intuitive, simple and achieve a FI score that is often consistent with other FI measures at a low computational cost. Unfortunately, they are also extremely biased towards categorical variables with large cardinalities \citep{strobl2007bias}, as later described in Section \ref{fi_bias}. 

Among the model agnostic global FI methods, the Permutation FI (PFI) \citep{breiman2001random}  is perhaps the most common approach. Its rationale is as follow. By randomly permuting the $j^{th}$ feature, its original association with the response variable is broken. Assume that the $j^{th}$ feature is associated with the outcome. As we apply the permuted feature, together with the remaining un-permuted features, to a given learning algorithm, the prediction accuracy is expected to substantially decrease. Thus, a reasonable measure for FI is the difference between the prediction error before and after permuting the variable.
The PFI measure relies on random permutations which greatly varies. Unfortunately, to achieve a reliable estimate, it is required to conduct multiple permutations, which is computationally demanding. 
PFI can be calculated both on the train-set and on the test-set, where each approach has its advantages \citep{molnar2020interpretable}. 
The train-set can be used to estimate the amount of model reliance on each feature. However, model error based on train data is often too optimistic and does not reflect the generalization performance \citep{molnar2020interpretable}. Using the PFI on the test-set reflects the extent to which a feature contributes to the model on unseen data, but may not reflect cases where the model heavily relies on a feature (see Section \ref{simulation_studies}). A key problem with PFI is in cases where the dataset contains correlated features. 
Here, forcing a specific value of a feature may be misleading. For example, consider the case of two given features, \textit{Gender} and \textit{Pregnant}. Forcing the gender variable to be \textit{Male} might result in a pregnant male as a data point. Further, it was shown that correlated features tend to "benefit" from the presence of each other and thus their PFI tends to inflate \citep{nicodemus2010behaviour}.

The leave one covariate out (LOCO) \citep{lei2018distribution} is another type of a model agnostic FI measure, which resembles in spirit to the PFI. LOCO evaluates the variable importance as the difference between the test-error of a model which was trained on all the data, and a model which is trained on all the data but the specified variable. Since training a model can be time consuming, this method may not be suitable in the presence of large data with many observations and features. 
Similarly to the PFI, LOCO may vary quite notably. Therefore, to obtain a reliable estimate, it is better to repeat this procedure a large number of times. Unfortunately, it substantially increases the execution time.  
Another popular global FI measure is the surrogate model. A global surrogate model trains an interpretable ML model in order to estimate the predictions of a black-box model. Then, the FI of the black-box model derived from the interpretable surrogate model. Finally, it is important to mention that global FI can also be obtained by averaging the results of local FI measure (see \ref{local_fi_measures} below) measures across observations. Specifically, a global FI of a feature is just the mean of the local FI for this variable, over all observations.

\subsubsection{Local FI Methods} \label{local_fi_measures}
Similarity to global FI, the local methods are also either model-specific or model-agnostic.  The model-specific measures  typically obtain a FI
score for each tree, and average the scores along the ensemble. \cite{sabaas} was the first to propose a local FI for decision tree ensembles. Inspired by Gain FI, \cite{sabaas} evaluates the change of the expected value before and after a split in non-terminal nodes. 

To attain a model-agnostic local FI measure it is possible to utilize a local surrogate model. For example, \textit{local surrogate model interpretable model-Agnostic} (LIME) \citep{ribeiro2016should} was proposed to attain a FI for a specific observation by training an interpretable model in the proximity of this observation. Specifically, for each observation, LIME creates an artificial dataset and trains an interpretable model on this data. The observation FI is estimated as the global FI of the interpretable model. In \textit{SHapley Additive exPlanations} (SHAP) \cite{lundberg2017unified} unify surrogate models with ideas from cooperative game-theory by estimating the \textit{Shapely} value.  
The \textit{Shapely} value of each prediction is the average marginal contribution of a feature to the prediction over all possible feature subsets. The Shapely value has many theoretical guarantees and desirable properties. In general, obtaining the exact \textit{Shapely} value in a model agnostic scenario requires exponential time. Later, \cite{lundberg2018consistent} introduced an exact polynomial-time algorithm to compute the SHAP values for trees (TreeSHAP) and tree ensembles.
%Since the SHAP values can be either positive or negative, Lundberg et al.\citep{lundberg2018consistent} propose to average the mean absolute SHAP value to attain a global FI. Further, Lundberg et al.\citep{lundberg2018consistent} show that existing global FI methods such as Gain and Sabbas \citep{sabaas}, may underestimate the FI in some cases. 
Although theoretically promising, TreeSHAP can assign a non-zero value to features that have no influence on the prediction \citep{molnar2020interpretable}.

Along the global and local FI methods, there are graphical approaches to further interpret the decision making principles of a black-
box ML algorithhm \citep{molnar2020interpretable}.

In this work, we focus on global FI measures since they are more popular, and demonstrate the bias of high cardinality categorical features most clearly and concisely. For the rest of the paper, and unless stated otherwise, global FI is referred to as FI.      

\subsection{Bias in Tree Based Algorithms} \label{fi_bias}
It is well-known that decision trees are biased towards categorical features with many categories. The reason for this phenomenon is quite obvious. During the training process, a node is split according to the variable which minimizes the error on the train-set (\ref{equation2}). Since categorical features allow more possible splits than numerical ones, they are more likely to be selected for a split. For example an \textit{ID} feature, which consists of a unique category for each observation, would minimize the spitting criterion compared to any possible feature. 
Based on this observation, most contributions which aim to achieve an unbiased FI measure focus on eliminating the bias on the tree level \citep{strobl2007bias, painsky2016cross}.

% Another approach to attain an unbiased FI is by P-values computed with Permutation Importance (PIMP) \citep{altmann2010Permutation}. 
% PIMP \citep{altmann2010Permutation}, a heuristic method to normalize a biased feature importance. First PIMP computes a prmutation FI on 
% PIMP compares and to retrain a model based on the highest scoring features of that approach.
% PIMP permutes the response variable multiple times and builds a model for each Permutation and for each model, the FI is estimated, called the null importances. The normalized FI is the p-value of the original FI under the distribution of the null importances, Gained by a maximum likelihood estimate of an assumed parametric distribution or by a non-parametric method. The advantage of this approach is that the relationship between predictor variables is unchanged. Except for the computational complexity of fitting multiple models, PIMP fixes only the FI estimates and not the algorithm itself; 

A variety of solutions to tree-bias were suggested over the years. The first line of work focuses on unbiased variable selection using hypothesis testing frameworks \citep{loh2002regression,kim2003classification,loh1997split}.
For example, in QUEST \citep{loh1997split}, the authors  
test the association of each feature with the labels and choose the variable with the most significant p-value to split. For numerical features, the p-values are derived from an ANOVA F-statistics while in categorical features they are derived from a $\chi^2$ test. Later, Hothorn et al. generalized these methods by conditional inference trees (CIT), which utilizes nonparametric tests for the same purpose  \citep{hothorn2006unbiased}. 
These approaches are built on a well defined statistical theory,
which either assumes a-priori modeling assumptions on the distribution of the features \citep{loh1997split} or using permutation or nonparametric tests such as in CIT \citep{hothorn2006unbiased}. 
One of the main problems with these methods are the a-priori assumptions, which are not always reasonable. On the other hand, nonparametric statistical tests may be computationally demanding. When an ensemble of trees is sequentially built  (as in GBM) this problem becomes more evident. Further, notice that the most ''informative" feature at each node is the one that splits the observations such that the generalization error (GE) is minimal. This is not  necessarily the variable that achieves the smallest p-values under the null assumption \citep{painsky2016cross}. Thus, these methods may decrease the predictive performance of CART.

For these reasons, a second line of work suggests ranking categorical features by their estimated GE. Ranking categorical features according to their GE was previously proposed for trees with k-way splits by \cite{sabato2008ranking} using Leave-One-Out (LOO) cross-validation \citep{frank1996selecting,frank1998using}. Recently, \cite{painsky2016cross} introduced \textit{adaptive LOO variable selection} (ALOOF) that is applicable for binary trees. ALOOF constructs an unbiased tree by conducting a ''fair" comparison between both numerical and categorical variables, using LOO estimates. This way, a large cardinality categorical feature is selected for a split if it proves to be effective on out-of-sample observations. 
%Our base tree differs from ALOOF only by using K-fold CV instead of LOO, which has better complexity guarantees in regression trees, as \citep{painsky2016cross} suggests. 

\section{Formulation Of CVB}
\label{Formulation Of CVB}
CVB introduces a single modification to the original GBM framework.
We hereby present the K-fold approach which is used in our CVB implementation. At the node selection phase, 
CVB selects the split based on cross-validation, and then splits it similarly to CART. 
Let $\mathcal{A}$ be the collection of all observations in a given node. We randomly divide the samples in $\mathcal{A}$ to $T$ equal sets of observations. Let $\{\mathcal{A}_t\}_{t=1}^T$ be the $T$ (non-overlapping) sets. For each set $\mathcal{A}_t$ we denote $\bar{\mathcal{A}}^t$ as the set of all observations that are not in $\mathcal{A}_t$. Specifically,  $\bar{\mathcal{A}}^t = \mathcal{A} \setminus \mathcal{A}_t$. For every $t=1,\dots,T$, we find the optimal split over $\bar{\mathcal{A}}^t$, according to (\ref{equation1}). Then, we evaluate the permanence of the split on $\mathcal{A}_t$. The rank of the feature is the average performance over all $t=1,\dots,T$. We choose the split with the lowest rank and split it following CART's splitting rule. Notice that CVB is similar in spirit to \citep{painsky2016cross}, as it applies a K-fold CV to evaluate every possible split.
A Python implementation of CVB is publicly available at Github\footnote{https://github.com/aba27059/unbiased\_fi\_for\_gb}. 

\subsection{CVB Stopping Criteria}
Tree algorithms like CART require a stopping criterion to avoid over-complex models that typically overfit. In GBM, it is most common to restrict the tree depth rather than using pruning approaches for speed considerations. Another common regularization technique in trees is the minimum gain decrease as shown in (\ref{equation2}). If the gain of the best split is less than a predefined hyper-parameter, the tree branch stops to grow and becomes a leaf. Although using the minimum gain decrease is a valid method to avoid over-fitting, it is more difficult for humans to interpret, as it is measured in units that depend on the impurity criterion and change greatly from one dataset to the other. In CVB, a tree ceases to grow if the best estimated GE across all features is greater than the impurity before the split. The result is a built-in 
regularization mechanism, which resembles minimum gain decrease. In practice, it leads to more interpretable models; CVB demonstrates a significant reduction in the number of trees, as well as a reduction in the number of leaves.

\section{Methods}
\label{Methods}
In the following sections we compare different FI measures in current GBM implementations on binary classification and regression tasks.
We use GBM implementations that provide built-in categorical variables handling. Specifically, CatBoost, LGBM, and Friedman's GBM (Vanilla GBM). We leave XGBoost outside the scope of this paper as it requires either one-hot-encoding which results in unfeasible run-time on the studied problems. 

% which we refer to as \textit{Vanilla GBM}

We use the following FI measures: Gain, PFI, and SHAP. Since SHAP is a local method, we derive its global FI\footnote{https://github.com/slundberg/shap}. 
It is important to emphasize that these FI measures are not (directly) integrated in some of the GBM implementations mentioned above. Specifically, CatBoost interface does not offer Gain FI nor node inspection data. Therefore, we use its Gain inspired default FI as in the official documentation\footnote{https://CatBoost.ai/docs/concepts/fstr.html}.
Further, since SHAP FI official package has built-in integration for CatBoost and LGBM open-source implementations, we report it when possible or refer to its results in the Appendix. 

We set the following hyper-parameters across all our experiments (and for all the studied algorithms): maximum tree depth = $3$, number of base learners (number of trees) = $100$, learning rate (shrinkage) = $0.1$. We do not change other default hyper-parameters values and do not apply bagging (stochastic GBM). We focus on mean squared error (MSE) to measure the regression error and log-loss \citep{painsky2018universality} to measure the binary classification error. When evaluating PFI, we apply twenty permutations to estimate the decrease in model performance. On real datasets, we use K-fold cross-validation to compute error and FI metrics (with K = 30 unless stated otherwise). In CVB, for the node selection phase, we use $T = 5$. Throughout our experiments, we report the scaled FI. That is, we scale the FI values over all variables, so they sum to one\footnote{In some cases, the PFI can also be negative; we consider this case as zero importance.}.

% \footnote{When CVB outputs a model with one leaf, we consider the FI as zero under the convention of $\frac{0}{0} = 0$.}

\section{Bias In Gradient Boosting FI} \label{simulation_studies}
We start with a simple synthetic data experiment that demonstrates the FI bias in current GBM implementations. We follow the setup introduced by \cite{strobl2007bias}, 
which studied a binary classification task with five features. In their setup, X0 is a numeric feature which follows a standard normal distribution, and X1 to X4 are four categorical features which follow a uniform distribution over alphabet sizes 10, 20, 50 and 100 respectively. In our experiment, we draw $n=6000$ observations, train a GBM model and compare the corresponding FI measures. We repeat this experiment $100$ times and report the average merits.
\subsection{Null Case}
In the first experiment, which we refer to as the \textit{null case}, the target variable is independent of the features and follows a Bernoulli distribution with a parameter $p=0.5$. In this setup, none of the features are informative with respect to the response variable. Thus, a perfect FI measure shall score each feature a zero score.

\begin{figure*}[t]
\centering
\fontsize{18}{28}
\textbf{Simulation 1: \textit{null case}}\par\medskip
\hspace*{-0.5cm} 
\includegraphics[width=13.5cm, height = 8cm]{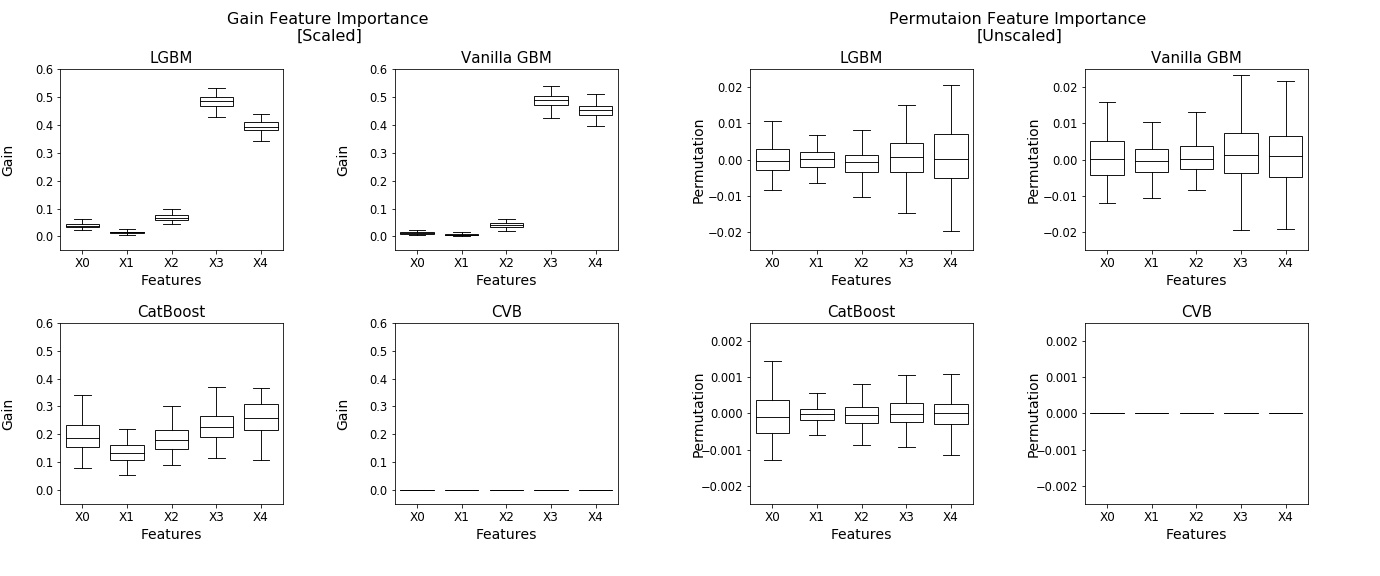}
\vspace*{-10mm}
\caption{\textit{Scaled Gain FI (left) and PFI (right) for the \textit{null case} experiment where all displayed features are uninformative.}}
\label{null_exp}
\end{figure*}

Figure \ref{null_exp} demonstrates the PFI and Gain FI results. The Gain FI \textit{(left)} illustrates the discussed bias in categorical variables. Specifically, we observe that categorical features with high cardinality tend to attain a greater FI than the rest, despite the fact that all features are not informative. In vanilla GBM and LGBM this phenomenon is more evident where $X3$ and $X4$ account together for almost $90\%$ of the overall FI whereas in CatBoost we observe a more subtle and monotone effect. CVB reliably recognizes uninformative features and scores zero importance for each feature. It is important to emphasize that although the reported scores are scaled, CVB attains zero FI scores. The reason lies on the standard convention, stating  that if all scores are zero defined (which corresponds to stub trees), then their scaled score is zero as well. 

The right charts in Figure \ref{null_exp} demonstrate the PFI.
Here, the mean FI of uninformative variables is around zero. This is not quite surprising, as the PFI is measured on the test-set. 
% Links that algorithms learned on the train data can not generalize to test data as the label is pure noise; 
Nevertheless, we observe that features with high cardinality result in a greater variance than the rest, both in LGBM, Vanilla GBM, and CatBoost. CatBoost outperforms LGBM and Vanilla GBM and obtains around 10 times smaller FI for the uninformative variables. CVB outperforms CatBoost and again scores the uninformative features with zero importance.
For LGBM and CatBoost, the SHAP FI (Appendix A) demonstrates the same monotonic behavior and scores similarly to the CatBoost Gain FI. Although in LGBM the slope is greater, CatBoost demonstrates a much greater variance.

\subsection{Power Case}
\begin{figure*}[t]
\centering
\fontsize{18}{28}
\textbf{Simulation 2: \textit{power case}}\par\medskip
\hspace*{-0.5cm}  
\includegraphics[width=13.5cm, height = 8.3cm]{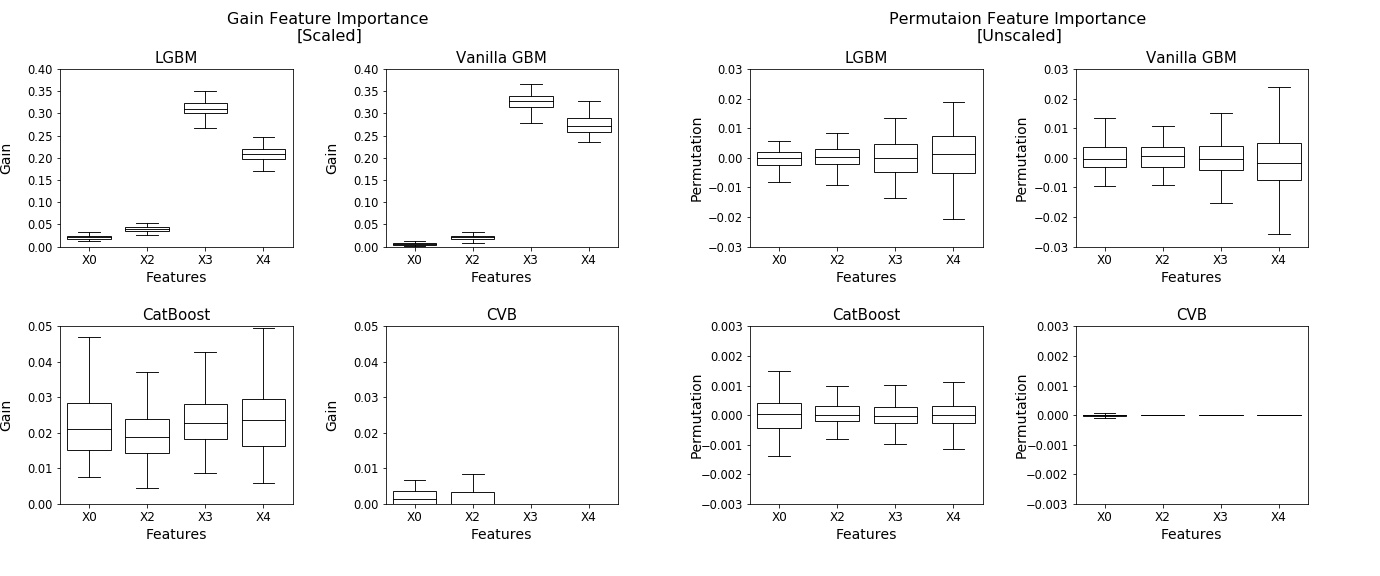}
\vspace*{-10mm}
\caption{\textit{Scaled Gain FI (left) and PFI (right) for the \textit{power case} experiment where only $X1$ is informative. Since $X1$ is informative it is on a different scale and it is threfore omitted to improve visualization. Its mean values are $0.186$, $0.181$, $0.166$, $0.145$ for Vanilla GBM, LGBM, CatBoost and CVB respectively.}}
\label{power_exp}
\end{figure*}

In the second experiment $Y$ depends on $X1$ in the following manner:
\begin{equation} 
  Y = \begin{cases}
      \text{Ber}(0.5 + \alpha) \text{,  X1} \in \{0,1,2,3,4\} 
        \\
        \text{Ber}(0.5 - \alpha) \text{,  X1} \in \{5,6,7,8,9\}
        \end{cases}
\end{equation}
where $\text{Ber}$ is a Bernoulli distribution and $\alpha$ is a predefined parameter. In this experiment, for $\alpha>0$, a perfect FI measure would assign a positive importance to $X1$ whereas the rest of the features shall attain zero importance. Figure \ref{power_exp} demonstrates the results we achieve for the uninformative features (all but $X1$) for the case where $\alpha = 0.2$. The Gain FI \textit{(left)} assigns only $37\%$ and $42\%$ FI to $X1$ for Vanilla GBM and LGBM where CatBoost performs better with $91\%$ FI for $X1$. CVB outperforms all the baseline schemes with $99.5\%$ FI assigned to the informative feature. As with the \textit{null case}, LGBM and Vanilla GBM over-fit the train-set in cases where high cardinality features are present, and attain a significantly greater FI for $X3$ and $X4$. In CatBoost, this phenomenon is considerably more subtle. 

Similarly to the \textit{null case}, the PFI \textit{(right)} variance increases with the category size, for all the algorithms besides CVB. CatBoost outperforms the baselines with a variance that is an order of magnitude smaller than LGBM and Vanilla LGBM. Let us now focus on the scaled PFI. LGBM and Vanilla GBM assigns $95.4\%$ importance for $X1$ whereas CatBoost assigns  $99.4\%$ for it. CVB outperforms CatBoost and assigns $99.9\%$ FI for $X1$. Finally, SHAP FI (Appendix A) scores $55.3\%$ and $90.0\%$ FI for X1 in LGBM and CatBoost respectively. 

Following Strobel's study \citep{strobl2007bias}, we conclude that all GBM implementations exhibits FI bias to some extent. Specifically, Vanilla GBM and LGBM perform the worst and are highly prone to over-fit large cardinality categorical features. CatBoost is one scale better then the former, attaining a significantly smaller bias than its alternatives. Finally, CVB is superior even to CatBoost, and demonstrates a reliable FI measure with almost no bias.

\section{Real Data Case Studies}
\label{Real Data Case Studies}
We now apply the CVB approach to real-world datasets and compare it to CatBoost, LGBM, and Vanilla GBM. We examine a collection of datasets (see Tables \ref{regression_error} and \ref{classification_error}) which typically consists of categorical features with a relatively large number of categories. As opposed to Section \ref{simulation_studies}, we do not know the ``true" FI values in these real-world problems. Therefore, we focus our attention to datasets for which the studied methods disagree. It is important to emphasize that the examined datasets undergo a preprocessing stage which eliminates allegedly uninformative features (such as indexing, or different group membership indicators). These features typically cause severe FI inaccuracies in currently known methods, as opposed to our proposed framework (as demonstrated in Section \ref{simulation_studies}). We do not report these results for brevity, and only focus on the more insightful examples. 
We start with a comprehensive case study of the Amazon dataset\footnote{https://www.kaggle.com/c/amazon-employee-access-challenge}.

% We now apply the CVB approach to real-world datasets and compare it with CatBoost, LGBM, and Vanilla GBM. \textbf{We examined various datasets (see Table \ref{regression_error} and \ref{classification_error}) and focus on the ones that display FI anomalies, mainly datasets that include categorical features with a relatively large number of categories}. We start with a comprehensive case
% study of the \textit{Amazon} dataset\footnote{https://www.kaggle.com/c/amazon-employee-access-challenge}.

\subsection{Amazon Dataset}
The \textit{Amazon} dataset is a binary classification task where the purpose is to predict whether a specific employee should get access to a specific resource. The dataset contains $n = 32,769$ observations and $p = 9$  categorical variables, including \textit{Resource} - an ID for each resource (7518 categories), and  \textit{Mgr\_id} - the manager ID of the employee (4243 categories).
 
\begin{figure*}[t]
 \centering
\fontsize{18}{28}
 \textbf{Amazon FI}\par\medskip
   \vspace*{-1.5mm}
\hspace*{-1.75cm}
\includegraphics[width= 15.5cm, height = 6.3cm]{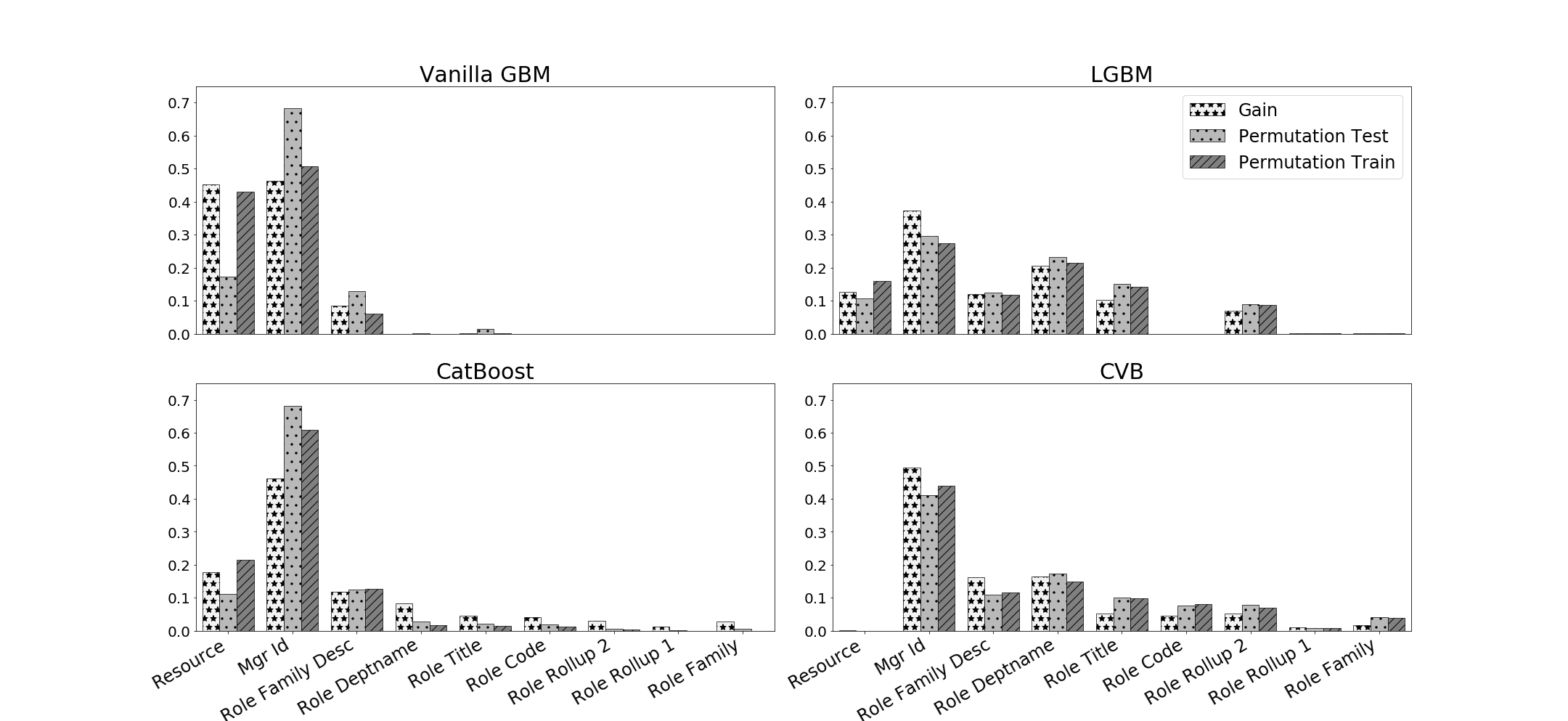}
\vspace*{-10mm}
\caption{\textit{Scaled FI on the  \textit{Amazon} dataset, across $30$ folds. Variables are ordered by their cardinality, from left to right, in descending order.}}
\label{amazon_fi}
 \vspace*{5mm}
\end{figure*}

We begin our analysis by applying different GBM algorithms and corresponding FI measures (Figure \ref{amazon_fi}).  The reported results are obtained by a 30-fold cross validation procedure, to attain statistically meaningful results. As we study the results we attain, we observe the following:
\begin{enumerate}

\item As in the simulation experiments, Vanilla GBM mostly utilizes high cardinality features and attains large values of FI 
for higher cardinality features (which appear on the left side of the plots).

\item  The \textit{Resource} variable attains a significantly large FI value in all methods besides CVB (in a significance level of 1\%).

\item  As we further study the \textit{Resource} variable, we observe that Vanilla GBM, LGBM, and CatBoost demonstrate a significant gap between the PFI on the train-set and the PFI on the test-set. Since both are in the same units, a gap in favor of the former suggests that this feature is quite dominant in the train-set, but has less effect on the test-set. This may suggest an over-fitted model.
\end{enumerate}

\begin{figure}[ht]
%\hspace*{-1cm} 
 \centering
\fontsize{14}{28}
%\hspace*{3mm}
\textbf{Error on Amazon}\par\medskip
%\vspace*{-2mm}
    %\includegraphics[width=7cm, height = 5cm ]{plots/error_on_amazon_with_without_Resource}
    \hspace*{-5mm}
    \includegraphics[width =0.5\textwidth,bb= 5 0 520 480,clip]{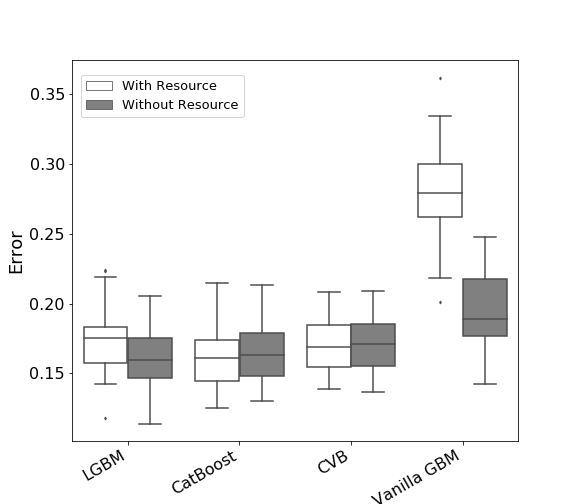}
%\vspace*{-2mm}
\caption{\textit{Error (log-loss) on the  \textit{Amazon} dataset, across 30 folds, with and without the \textit{Resource} variable.}}
\label{amazon_error}
\end{figure}

To validate the alleged over-fitting, we conduct the following experiment.  We compare the prediction error of each GBM implementation with and without the \textit{Resource} variable. Figure \ref{amazon_error} illustrates the results we achieve. We observe that Vanilla GBM and LGBM perform better without the \textit{Resource} variable. CatBoost and CVB demonstrate quite similar results - while the median error in CatBoost is smaller than CVB, its variance is greater.
In both cases it is not clear whether the \textit{Resource} variable contributes to the model performance. However, CVB is the only method that reflects it in its FI.

Overall, we conclude that some implementations may rely on features in a different manner. For example, CVB does not rely on the \textit{Resource} variable while CatBoost does depend on it and may even benefit to its prediction accuracy.  
Finally, we observe that  CVB FI methods are consistent and equal to each other. This means that even simple approaches as Gain FI perform well without the need for more complex FI methods.

\subsection{Criteo CTR Dataset}
The \textit{Criteo Click-Through Rate} (CTR) challenge\footnote{https://www.kaggle.com/c/criteo-display-ad-challenge} is a binary classification task where the goal is to predict whether a user clicks on an on-line ad. The dataset contains $n = 40,428,967$ observations and $p = 23$ categorical variables. Due to run-time considerations, we decrease the size of the dataset in the following manner. We randomly sample $30,000$ observations, such that half of them have a positive label and the others are negative (following \cite{he2009learning}). We remove the \textit{ID} feature which is unique identifier of each observation. We remain with $p = 22$ variables including:
\textit{Device\_ip} (25,573 categories), \textit{Device\_id} (4,928 categories), \textit{Device\_model} (2,154 categories). We apply different GBM methods and obtain the corresponding FI measures. Figure \ref{criteo_fi} summarizes the results we obtain. We observe the following:

\begin{figure*}[t]
 \centering
\fontsize{18}{28}
 \textbf{Criteo CTR FI}\par\medskip
  \vspace*{-1.5mm}
\hspace*{-1.75cm}
\includegraphics[width= 15.5cm, height = 6.3cm]{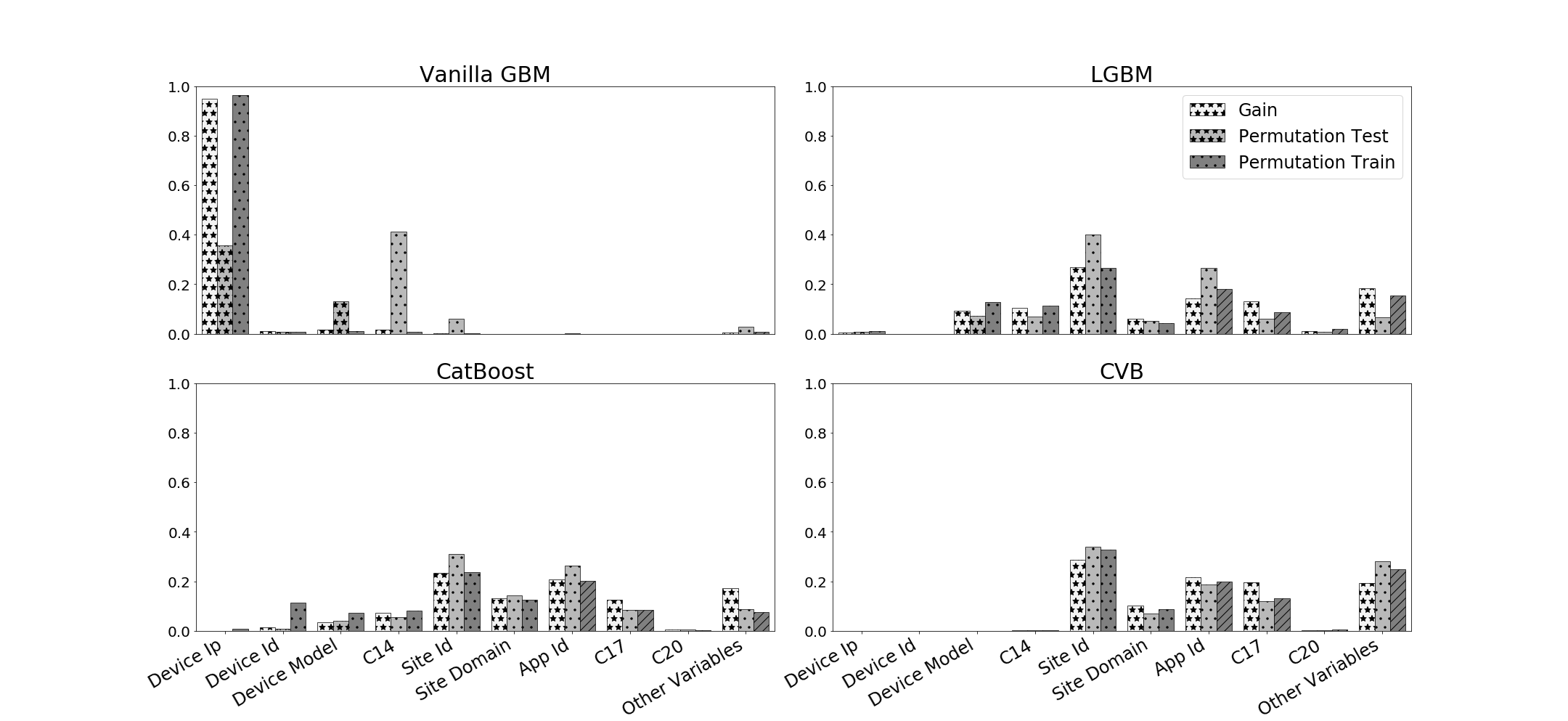}
\vspace*{-7mm}
\caption{\textit{Scaled FI on the \textit{Criteo CTR} dataset, across $30$ folds. Variables are ordered by their cardinality, from left to right, in descending order. For visualisation, FI results for features with low cardinality that have small FI values are grouped.}}
\label{criteo_fi}
\end{figure*} 

\begin{figure}[ht]
%\hspace*{-1cm} 
 \centering
\fontsize{14}{28}
%\hspace*{3mm}
\textbf{Error on Criteo CTR}\par\medskip
%\vspace*{-2mm}
\centering
    %\includegraphics[width=7cm, height = 5cm ]{plots/Error on Criteo_with_without.png}
%\vspace*{-2mm}
\hspace*{-5mm}
\includegraphics[width =0.5\textwidth,bb= 5 0 520 480,clip]{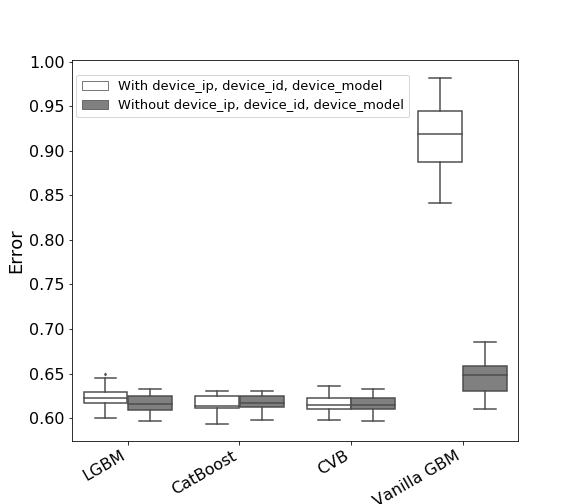}
\caption{\textit{Error (log-loss) on the \textit{Criteo dataset}, across 30 folds, with and without the following variables: \textit{Device\_ip, Device\_id, Device\_model}.}}
\label{criteo_error}
\end{figure}

\begin{enumerate}

\item As in the previous experiment, Vanilla GBM is biased towards high categorical variables and over-fits the train-set with Gain and PFI on train data close to one for the \textit{Device\_ip} feature. 

\item The results for LGBM and CatBoost are quite similar and score high cardinality features with larger scores compared to CVB results. They also over-estimate high cardinality features with a relatively small difference between the PFI on train data and PFI on test data for high cardinality features. 

\item In LGBM, \textit{Device\_ip} attains a large SHAP value in contrast to other FI measures while in CatBoost SHAP FI is more consistent with other metrics, especially Gain and PFI on test data. As in the previous experiment, we examine whether CVB FI results are reliable by comparing the models error with and without a feature set. 
Figure \ref{criteo_error} demonstrates the results we achieve by comparing 
the set of all features to a the set of features CVB scores a positive FI. The results are quite similar to the  \textit{Amazon} data-set experiment. 
\end{enumerate}

\subsection{Prediction Accuracy}
Open-source implementations such as CatBoost and LGBM have many hyper-parameters that are tuned to obtain the best results in practice. Further, some implementations introduce different enhancements to the original GBM implementation to achieve a even better prediction accuracy as discussed in Section \ref{gbm_implementations}. Since improving the prediction accuracy is not the main focus of this study, we show that CVB is competitive with LGBM, CatBoost, and Vanilla GBM in a fixed hyper-parameters setting (number of trees = 100, learning rate = 0.1, maximum tree depth = 3). Table \ref{regression_error} summarizes the 10-fold mean RMSE of each algorithm on four different well-known datasets - \textit{Allstate claim severity}, \textit{Bike rentals}, \textit{Boston house pricing} and \textit{Kaggle house pricing}
% \footnote{The datasets properties are described in the Appendix.}
. %In this setting, LGBM and Vanilla GBM perform the best, while CVB and CatBoost perform the worst. 
Table \ref{classification_error} demonstrates the classification error over six classification datasets - \textit{Amazon} \textit{Breast cancer}, \textit{Don't get kicked}, \textit{Criteo CTR}, \textit{KDD upselling} and \textit{Adult}. %In this setting, CatBoost demonstrates superior performance on average followed by CVB and LGBM. Vanilla LGBM is the worst by a large margin. 
In all of our experiments, CVB demonstrates competitive (and sometimes even superior) prediction accuracy to state-of-the-art methods.

\begin{table}
\centering
\fontsize{12}{28}
 \textbf{Regression RMSE}\par\medskip
%   \vspace*{-1mm}
\fontsize{9}{28}
\begin{tabular}{|p{2.6cm}|p{1.5cm}|p{1.5cm}|p{1.5cm}|p{1.5cm}|} \hline
Dataset & CVB & CatBoost & LGBM & Vanilla GBM\\ \hline
 \textit{Allstate}$^9$  & 0.3115  &  0.3242 &  0.3120 & 0.3123  \\ 
 &  (0.0036) &   (0.0036) &  (0.0038) &   (0.0041)\\ \hline

\textit{Bike Rentals}$^{10}$   &  0.2554  &  0.4484  &  0.2376  &  0.2470 \\ 
&  (0.021) &  (0.0415) &  (0.0096) &  (0.0173)\\ \hline

\textit{Boston HP}$^9$ & 0.0276  &  0.0241 &  0.0238 &  0.0243 \\
& (0.0143) &  (0.0102) &  (0.0108) &  (0.0115)\\ \hline  

\textit{Kaggle HP}$^9$ &  0.0185  &  0.0187 &   0.0162 & 0.0159 \\ 
& (0.0038) &  (0.0042) &  (0.0034) &  (0.0034)\\
\hline\end{tabular}
\vspace*{1mm}
\caption{\textit{Mean and standard deviation of the RMSE across 10 folds. For visibility reasons we applied log transformation on the target. HP refers to \textit{house pricing}.}}
\label{regression_error}
%\vspace*{-10mm}
\end{table}

\vspace*{1cm}

\begin{table}[ht]
\centering
\fontsize{12}{28}
 \textbf{Classification log-loss}\par\medskip
%   \vspace*{-1mm}
\fontsize{9}{28}
\begin{tabular}{|p{2.6cm}|p{1.5cm}|p{1.5cm}|p{1.5cm}|p{1.5cm}|} \hline
Dataset&{CVB}&{CatBoost}&{LGBM}&{Vanilla GBM}\\ \hline
KDD upselling$^8$&   0.1686 &  0.1681 &  0.1733 &  0.3416 \\ 
 &  (0.0074) &  (0.0072) &  (0.0071) &  (0.0145) \\ \hline

 \textit{Amazon}$^9$ &  0.1716 & 0.1606 &  0.1724 &  0.2795 \\ 
 &  (0.0186) &  (0.0216) &  (0.0243) &  (0.0355)\\ \hline

 \textit{Breast Cancer}$^{10}$ & 0.1091 & 0.0933 &  0.1080 &  0.1043\\ 
 &(0.0872) &  (0.0682) &  (0.1008) &  (0.1008)\\ \hline

 \textit{Don't Get Kicked}$^9$ &  0.3425 &  0.3433 & 0.3416 &  0.3584 \\ 
 & (0.0064) &  (0.0066) &  (0.0071) &  (0.0070)\\ \hline
 
 \textit{Criteo CTR} &  0.6161 &  0.6157 &  0.6241 &  0.9150 \\ 
 &  (0.0083) &  (0.0095) &  (0.0124) &  (0.0376) \\ \hline
 
 \textit{Adult}$^{10}$ &  0.2982 & 0.3021 & 0.2892 & 0.2917 \\ 
 &  (0.0043) &  (0.0095) &  (0.0088) &  (0.0038) \\ 
\hline\end{tabular}
\vspace*{0.1mm}
\caption{\textit{Mean and standard deviation of the log loss across 10 folds (30 on \textit{Amazon} and \textit{Criteo CTR}).}}
\label{classification_error}
%\vspace*{-7mm}
\end{table}

\newcommand\blfootnote[1]{%
  \begingroup
  \renewcommand\thefootnote{}\footnote{#1}%
  \addtocounter{footnote}{-1}%
  \endgroup
}

\blfootnote{$^8$https://www.kdd.org/kdd-cup/view/kdd-cup-2009/Data}
\blfootnote{$^9$https://www.kaggle.com}
\blfootnote{$^{10}$https://archive.ics.uci.edu/ml/datasets}

\section{Discussion And Conclusion}
\label{discussion}
Although common implementations of GBM utilize biased decision trees, they typically perform quite well, and demonstrate high prediction accuracy. Unfortunately, their FI is shown to be biased. To overcome this basic limitation, we introduce a CVB framework which utilizes unbiased decision trees. We show that CVB attains unbiased FI while maintaining a competitive level of generalization abilities. Further, we show that FI is not model-agnostic or universal. In fact, it is a unique property of each GBM implementation; given a prediction task, two different implementations may introduce relatively the same error but much different FI scores. 

CVB is a naive, not optimized implementation of GBM. Our experiments show that even this simple implementation results in better FI and a competitive accuracy. One may wonder how an optimized version of CVB may perform. For example, it is interesting to examine more efficient validation schemes, and introduce state-of-the-art GBM enhancements to our current CVB implementation. For example, a future enhancement of CVB may consider stochastic GBM with feature selection using out-of-bag examples. Finally, CVB may also be applied to correct the bias in local FI measures. This would result in accurate local FI measure for personalization purposes. 

% As well, we would like to examine further how the inner regularization scheme in ALOOF affects CVB proprieties. In this work, we used this mechanism as is in CVB, this has the advantage of producing significantly fewer trees and may be beneficial in low Resource environments. Despite this fact, it may over regularize CVB which is already restricted to a maximum tree depth of three as in the other methods we examined. Therefore, we propose to inspect if CVB which uses cross-validation only for feature selection without limiting the tree structure compares to this naive implementation.

% Since we have seen that CVB global FI is stable between different FI approaches and that the SHAP FI is highly correlated to other FI methods in CatBoost and LGBM, we believe that it will show the same FI as other methods in CVB empirically. Nevertheless, an interesting line of work can be to integrate SHAP or other local FI methods into CVB and to show that empirically. Thus, acquiring an accurate local FI which can be used for personalization purposes.

\bibliography{bibi}
\newpage
\section*{Appendix A}
\begin{figure}[ht]
\fontsize{12}{28}
\vspace*{-30mm}
\centering
% \fontsize{12}{28}
%  \textbf{SHAP FI - Simulations}\par\medskip
% \hspace*{-3cm}
%\includegraphics[width=8cm, height = 4.3cm ]{plots/Shap null.png}
\includegraphics[width =0.9\textwidth,bb= 20 -10 650 540,clip]{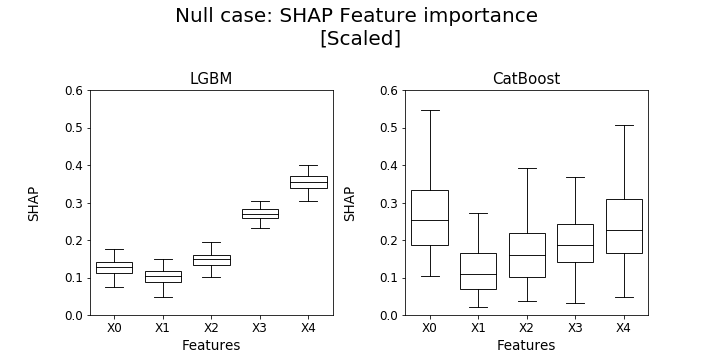}
\centering
\label{SHAP_null_case}
\vspace*{-4mm}
\caption{\textit{SHAP FI for the \textit{null case}.}}
% \vspace*{4mm}
\end{figure}

\begin{figure}[ht]
\vspace*{-3cm}
\centering
% \fontsize{12}{28}
%  \textbf{SHAP FI - Simulations}\par\medskip
  \hspace*{-5mm}
\includegraphics[width =0.9\textwidth,bb= 20 -10 650 540,clip]{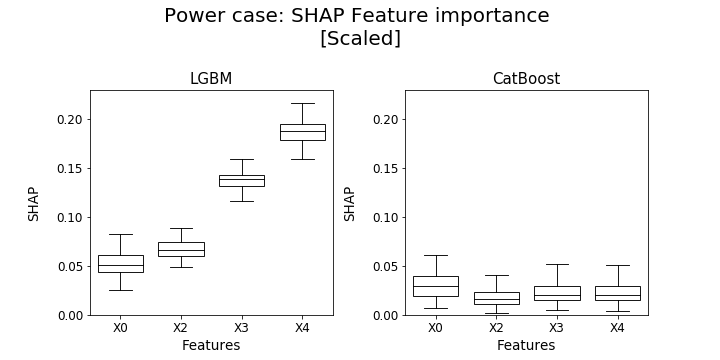}
\centering
\label{shap_power_case}
\vspace*{-4mm}
\caption{\textit{SHAP FI for the \textit{power case}.}}
% \vspace*{4mm}
\end{figure}

\begin{figure}
    \vspace*{-8cm}
    \centering
    
    \begin{minipage}{0.5\textwidth}
        \centering
        \textbf{SHAP FI - Amazon}\par\medskip
        \centering
        \includegraphics[width =0.9\textwidth,bb= 20 -20 900 950,clip]{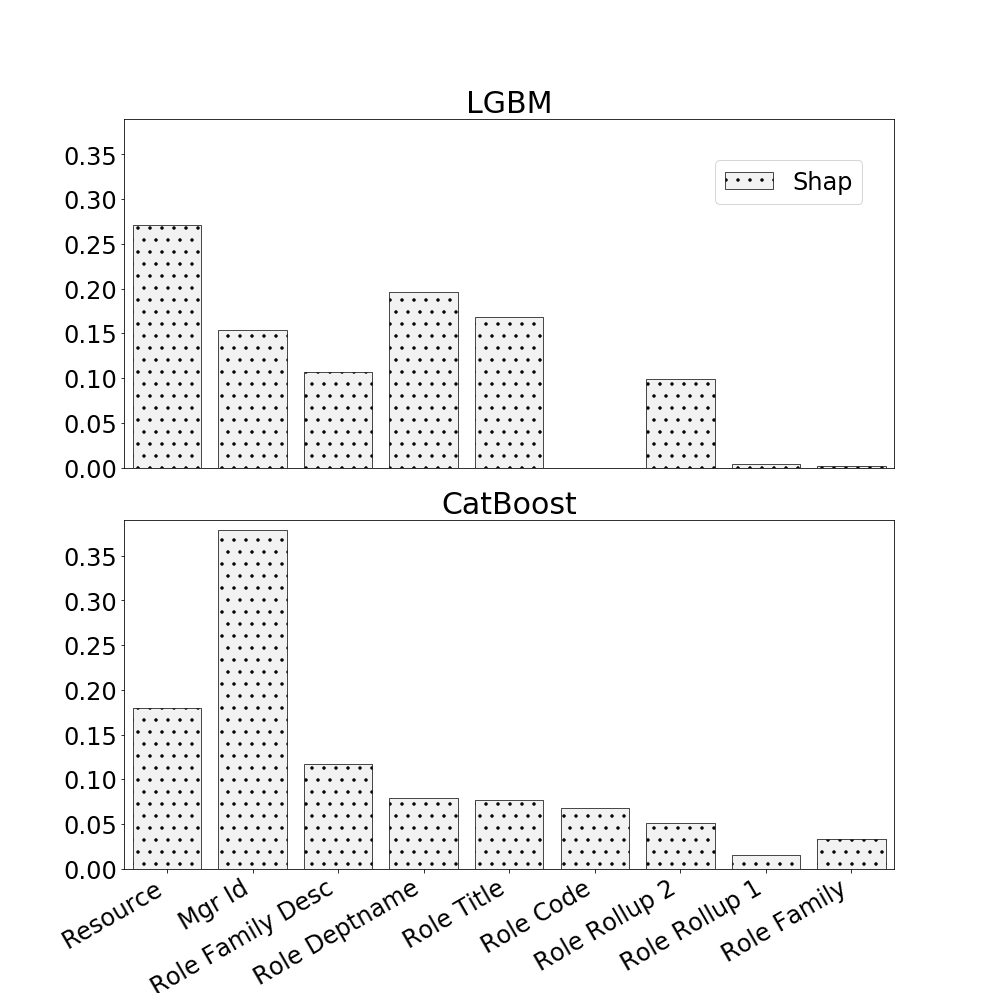}
        \label{amazon_fi_shap}
    \end{minipage}\hfill
    \begin{minipage}{0.5\textwidth}
        \centering
        \textbf{SHAP FI - Criteo CTR}\par\medskip
        \includegraphics[width =0.9\textwidth,bb= 20 -20 900 950,clip]{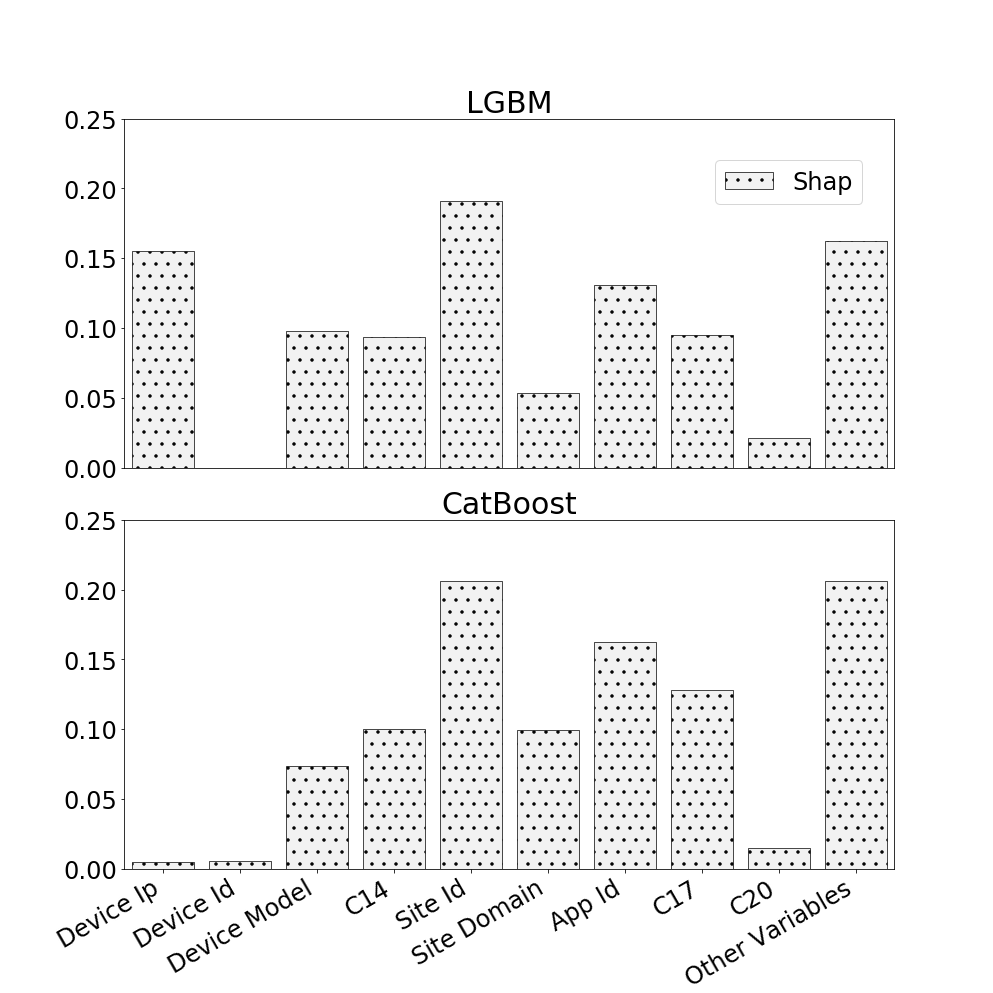}
        \label{criteo_fi_shap}
    \end{minipage}\hfill
    
\caption{\textit{FI on the \textit{Amazon and Criteo CTR datasets},including SHAP FI}}
\end{figure}

\end{document}